\documentclass[10pt,twocolumn,letterpaper]{article}

\usepackage{cvpr}
\usepackage{times}
\usepackage{epsfig}

\usepackage{graphicx}
\usepackage{amsmath}
\usepackage{amssymb}
\usepackage{bm}
\usepackage{subcaption}
\usepackage{multirow}
\usepackage[framed,numbered]{mcode}
\usepackage{algorithm}
\usepackage{algorithmic}
\usepackage{enumerate}
\usepackage{hyperref}
\hypersetup{pdfborder={0 0 0}, colorlinks=true,linkcolor=red}

\newtheorem{theo}{Theorem}
\newtheorem{lem}{Lemma}
\newtheorem{defi}{Definition}

\DeclareMathOperator*{\Diag}{Diag}
\DeclareMathOperator*{\X}{\mathbf{X}}
\DeclareMathOperator*{\M}{\mathbf{M}}
\DeclareMathOperator*{\Y}{\mathbf{Y}}
\DeclareMathOperator*{\x}{\mathbf{x}}
\DeclareMathOperator*{\y}{\mathbf{y}}

\cvprfinalcopy 
\def\cvprPaperID{****} 

\begin{document}

\title{Generalized Nonconvex Nonsmooth Low-Rank Minimization}

\author{Canyi Lu$^1$, Jinhui Tang$^2$, Shuicheng Yan$^1$, Zhouchen Lin$^{3,}$\thanks{Corresponding author.}\\
$^1$ Department of Electrical and Computer Engineering, National University of Singapore\\
$^2$ School of Computer Science, Nanjing University of Science and Technology\\
$^3$ Key Laboratory of Machine Perception (MOE), School of EECS, Peking University\\
{\tt\small canyilu@gmail.com, jinhuitang@mail.njust.edu.cn, eleyans@nus.edu.sg, zlin@pku.edu.cn}
}


\maketitle

\begin{abstract}
As surrogate functions of $L_0$-norm, many nonconvex penalty functions have been proposed to enhance the sparse vector recovery. It is easy to extend these nonconvex penalty functions on singular values of a matrix to enhance low-rank matrix recovery. However, different from convex optimization, solving the nonconvex low-rank minimization problem is much more challenging than the nonconvex sparse minimization problem. We observe that all the existing nonconvex penalty functions are concave and monotonically increasing on $[0,\infty)$. Thus their gradients 
 are decreasing functions. Based on this property, we propose an Iteratively Reweighted Nuclear Norm (IRNN) algorithm to solve the nonconvex nonsmooth low-rank minimization problem. IRNN iteratively solves a Weighted Singular Value Thresholding (WSVT) problem. By setting the weight vector as the gradient of the concave penalty function, the WSVT problem has a closed form solution. 
 In theory, we prove that IRNN decreases the objective function value monotonically, and any limit point is a stationary point. Extensive experiments on both synthetic data and real images demonstrate that IRNN enhances the low-rank matrix recovery compared with state-of-the-art convex algorithms.
\end{abstract}

\begin{figure}
	\begin{subfigure}[b]{0.235\textwidth}
		\centering
        \includegraphics[width=\textwidth]{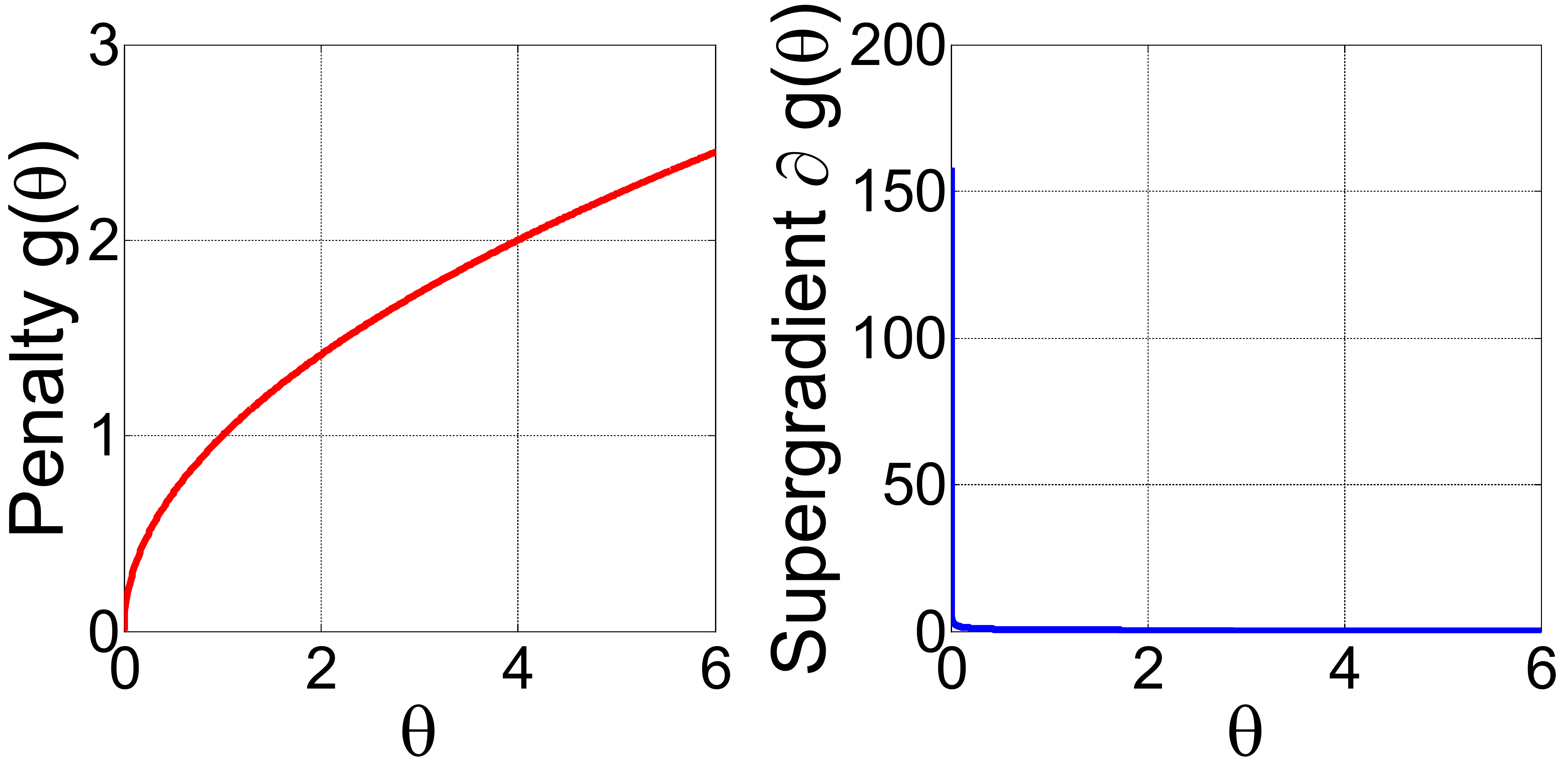}
        \caption{\scriptsize{$L_p$ Penalty \cite{frank1993statistical}}}
    \end{subfigure}
        \begin{subfigure}[b]{0.235\textwidth}
		\centering
		\includegraphics[width=\textwidth]{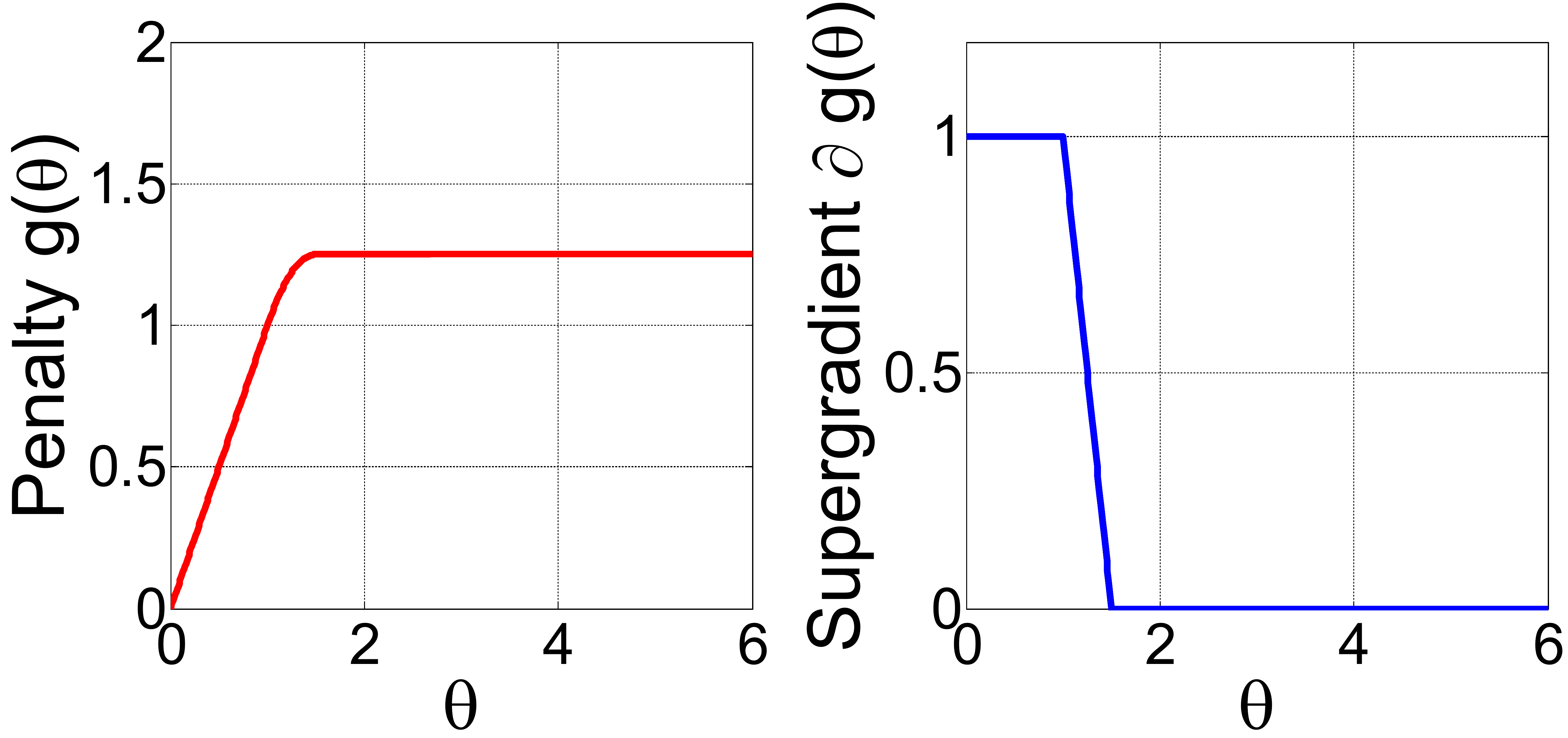}
		\caption{\scriptsize{SCAD Penalty \cite{fan2001variable}}}
    \end{subfigure}
	\begin{subfigure}[b]{0.235\textwidth}
		\centering
        \includegraphics[width=\textwidth]{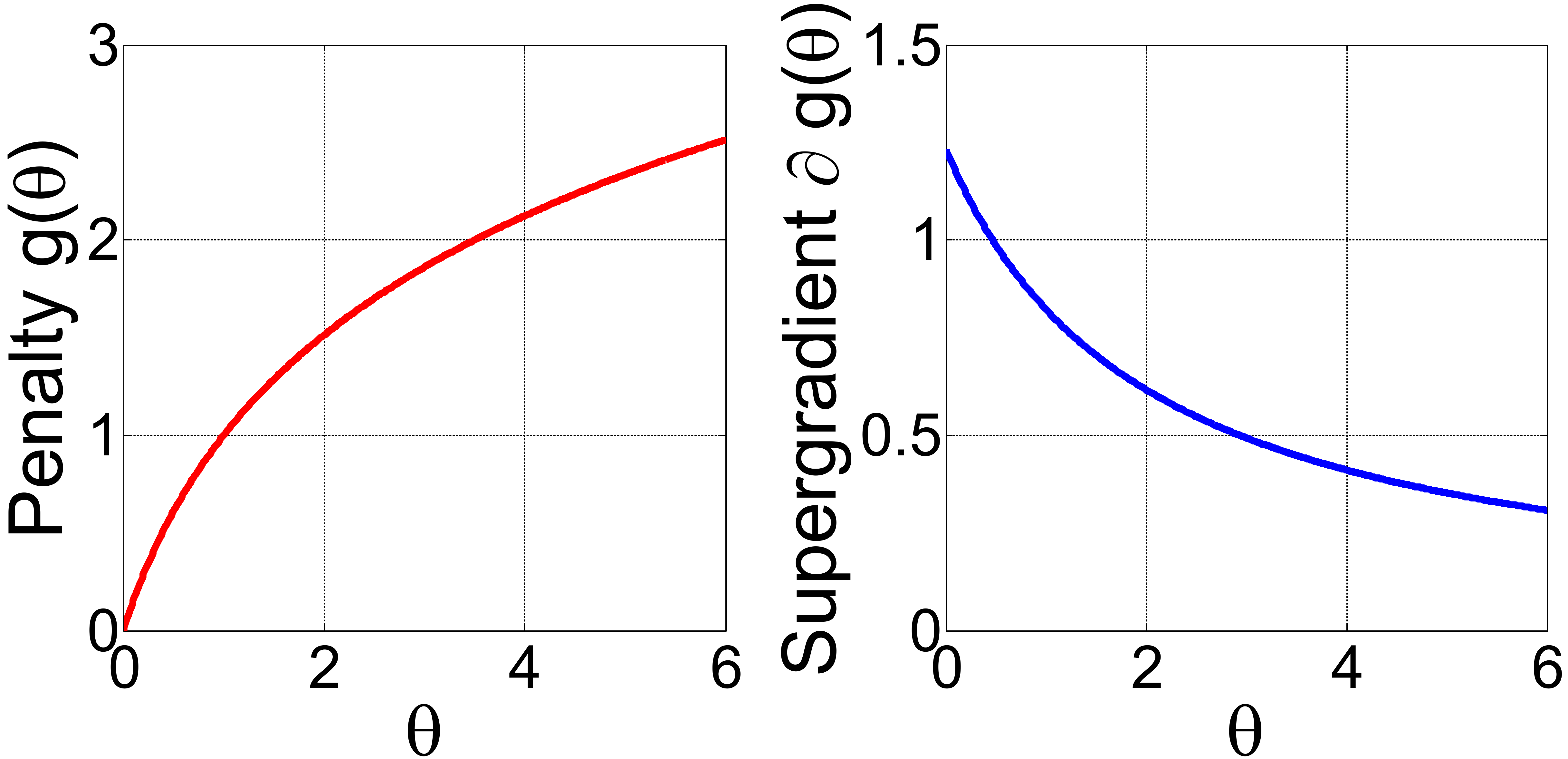}
        \caption{\scriptsize{Logarithm Penalty \cite{friedman2012fast}}}
    \end{subfigure}
        \begin{subfigure}[b]{0.235\textwidth}
		\centering
		\includegraphics[width=\textwidth]{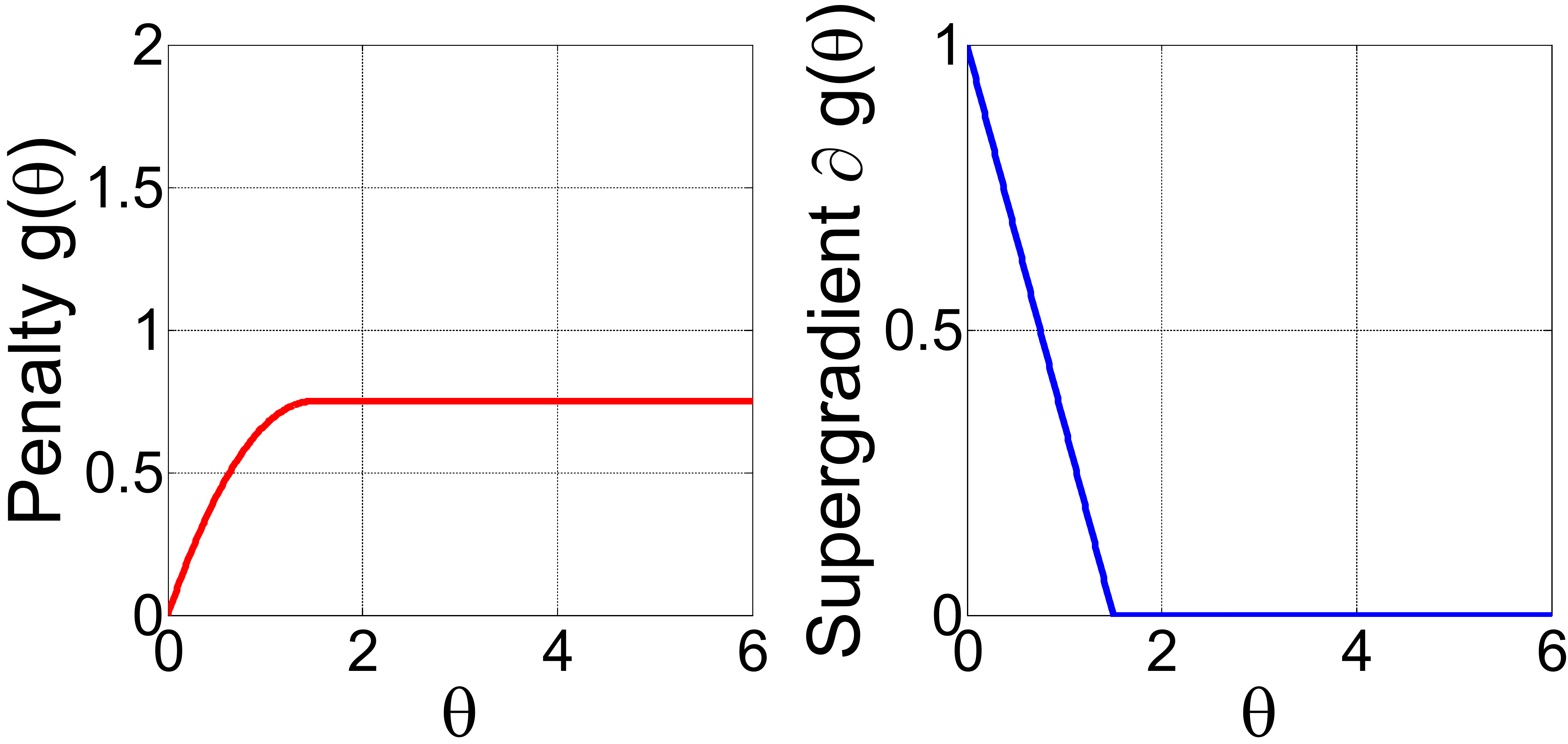}
		\caption{\scriptsize{MCP Penalty \cite{zhang2010nearly}}}
    \end{subfigure}
	\begin{subfigure}[b]{0.235\textwidth}
		\centering
        \includegraphics[width=\textwidth]{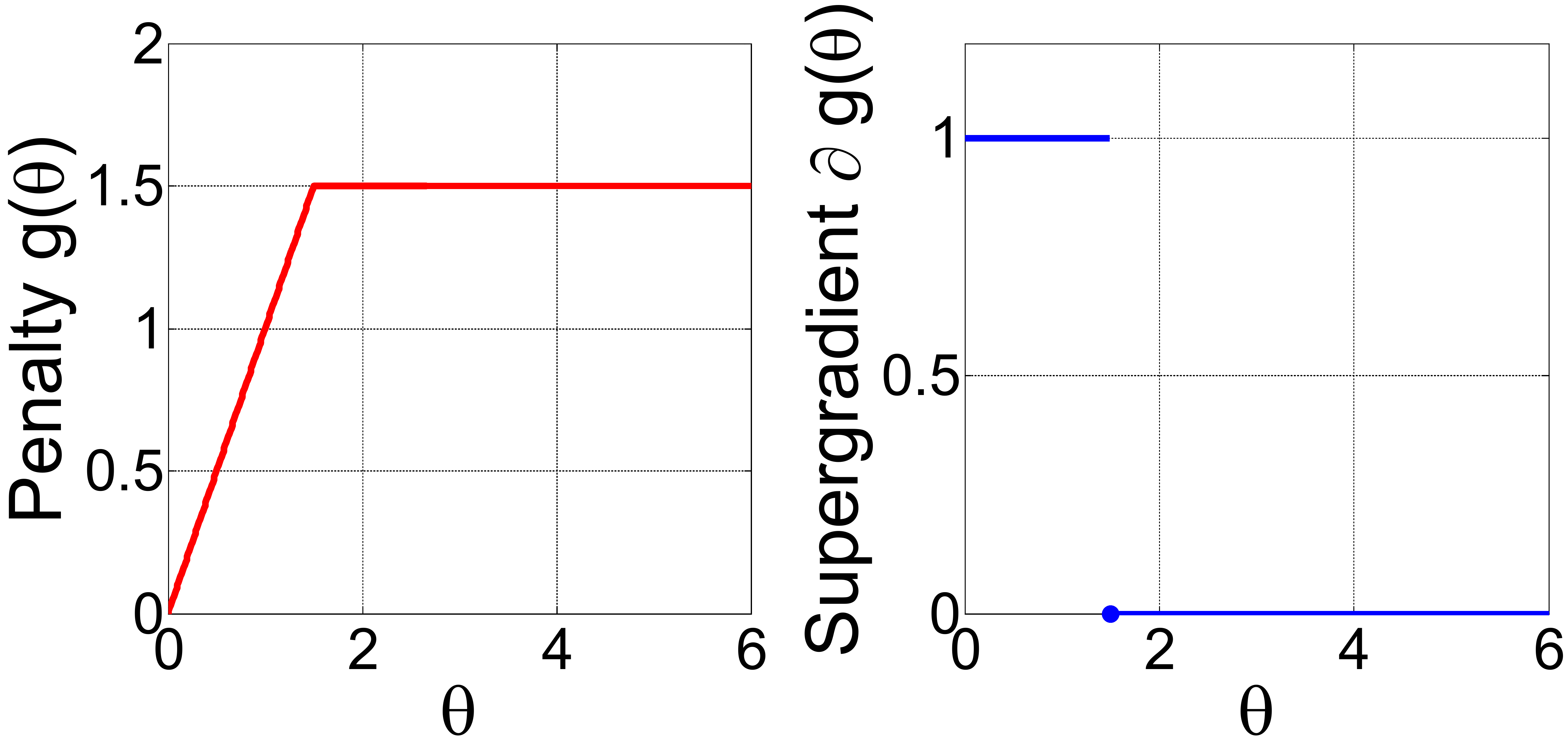}
        \caption{\scriptsize{Capped $L_1$ Penalty \cite{zhang2010analysis}}}
    \end{subfigure}
    \begin{subfigure}[b]{0.235\textwidth}
		\centering
		\includegraphics[width=\textwidth]{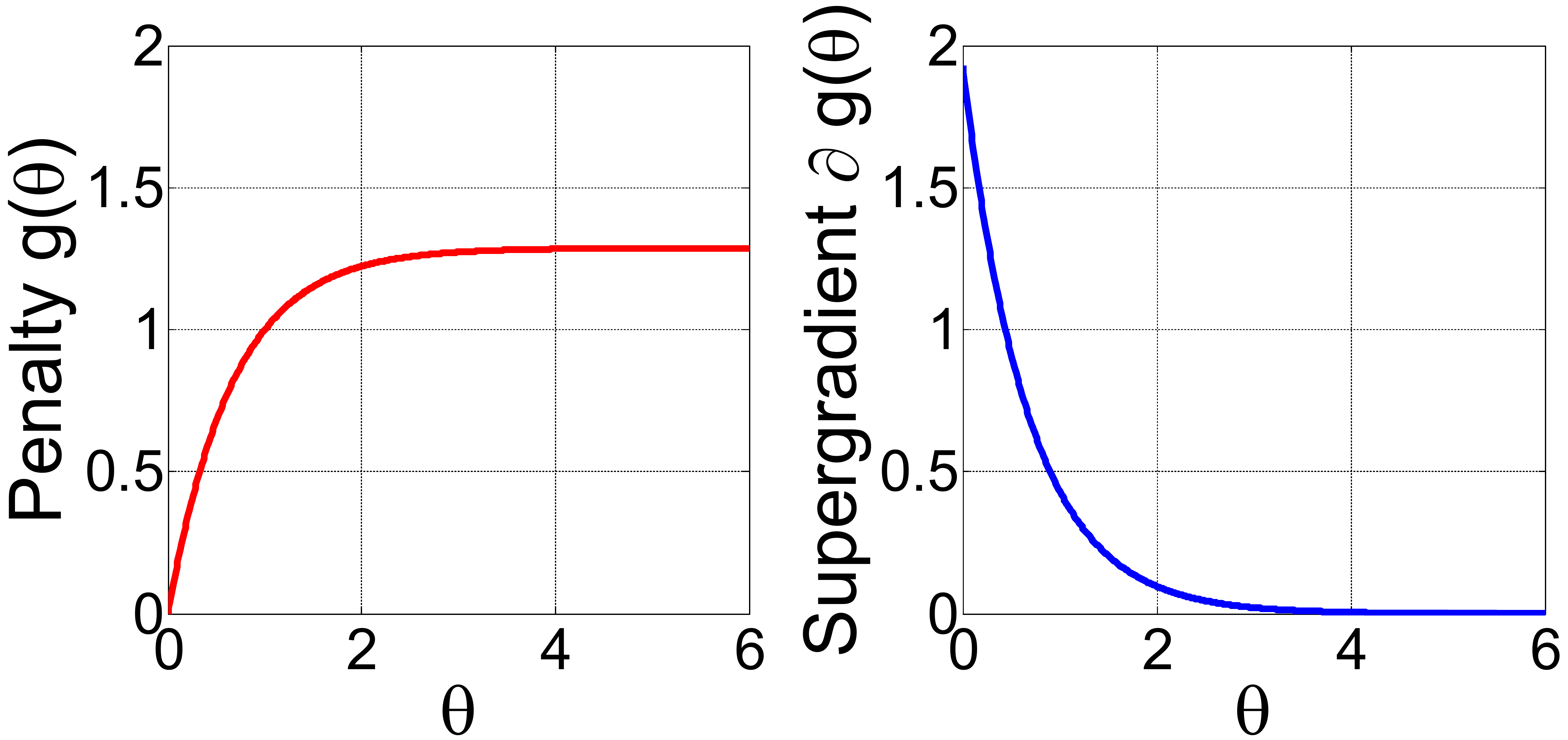}
		\caption{\scriptsize{ETP Penalty \cite{gao2011feasible}}}
    \end{subfigure}
    \begin{subfigure}[b]{0.235\textwidth}
		\centering
		\includegraphics[width=\textwidth]{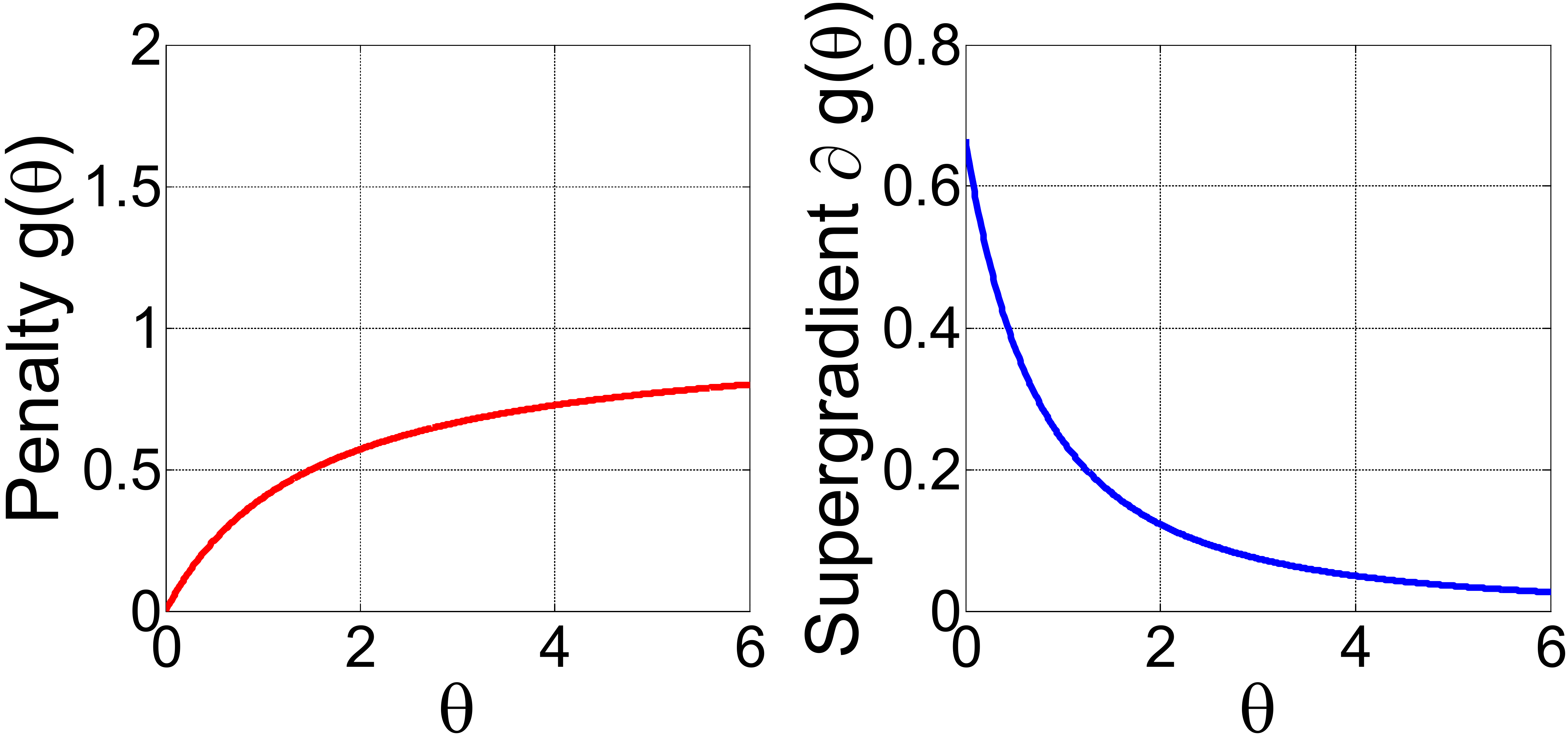}
		\caption{\scriptsize{Geman Penalty \cite{geman1995nonlinear}}}
    \end{subfigure}
    \begin{subfigure}[b]{0.235\textwidth}
		\centering
		\includegraphics[width=\textwidth]{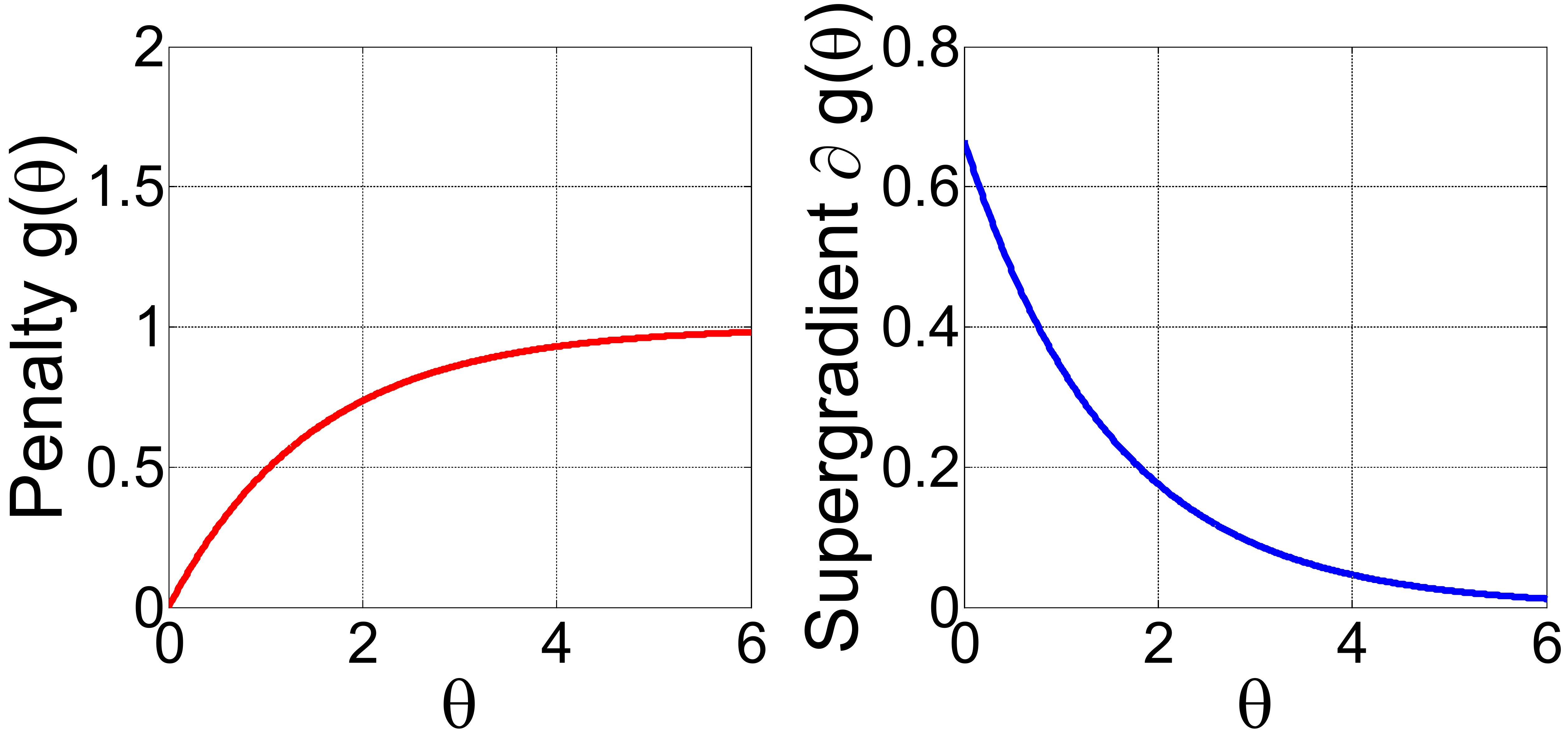}
		\caption{\scriptsize{Laplace Penalty \cite{trzasko2009highly}}}
    \end{subfigure}
\caption{\small{Illustration of the popular nonconvex surrogate functions of $||\theta||_0$ (left), and their supergradients (right). All these penalty functions share the common properties: concave and monotonically increasing on $[0,\infty)$. Thus their supergradients (see Section \ref{sec_21}) are nonnegative and monotonically decreasing. Our proposed general solver is based on this key observation.}}\label{fig_nonconfun}
\vspace{-1em}
\end{figure}

\begin{table*}[!t]
\scriptsize
\centering
\caption{\small{Popular nonconvex surrogate functions of $||\theta||_0$ and their supergradients.}}
\label{tab_nonpenlty}
\centering
\begin{tabular}{r| l| l }
\hline
Penalty & Formula $g_{\lambda}(\theta)$, $\theta\geq0$, $\lambda>0$ & Supergradient $\partial g_{\lambda}(\theta)$\\ \hline
$L_p$ \cite{frank1993statistical} & $\lambda\theta^p$
&  $\begin{cases}
		\infty, &	\text{ if } \theta=0,\\
		\lambda p\theta^{p-1}, & \text{ if } \theta>0.
		\end{cases}$\\\hline
SCAD \cite{fan2001variable} & $\begin{cases}
			 \lambda\theta, & \text{ if } \theta\leq\lambda, \\
			 \frac{-\theta^2+2\gamma\lambda\theta-\lambda^2}{2(\gamma-1)}, & \text{ if } \lambda<\theta\leq\gamma\lambda, \\
			\frac{\lambda^2(\gamma+1)}{2}, &\text{ if }\theta>\gamma\lambda.
			\end{cases}$
	&		$\begin{cases}
			 \lambda, & \text{ if } \theta\leq\lambda, \\
			 \frac{\gamma\lambda-\theta}{\gamma-1}, & \text{ if } \lambda<\theta\leq\gamma\lambda, \\
			0, &\text{ if }\theta>\gamma\lambda.
			\end{cases}$ \\\hline
Logarithm \cite{friedman2012fast}& $\frac{\lambda}{\log(\gamma+1)}\log(\gamma\theta+1)$ & $\frac{\gamma\lambda}{(\gamma\theta+1)\log(\gamma+1)}$ \\\hline
MCP \cite{zhang2010nearly}& $\begin{cases}
			 \lambda\theta-\frac{\theta^2}{2\gamma}, & \text{ if } \theta<\gamma\lambda, \\
			 \frac{1}{2}\gamma\lambda^2, & \text{ if } \theta\geq\gamma\lambda.
			\end{cases}$
	& $\begin{cases}
		\lambda-\frac{\theta}{\gamma}, &	\text{ if } \theta<\gamma\lambda,\\
		0, & \text{ if } \theta\geq\gamma\lambda.
		\end{cases}$\\\hline	
Capped $L_1$ \cite{zhang2010analysis}& 	
$\begin{cases}
	\lambda\theta, &	\text{ if } \theta<\gamma,\\	
	\lambda\gamma, & \text{ if } \theta\geq\gamma.
\end{cases}$	
& 	
$\begin{cases}
	\lambda, &	\text{ if } \theta<\gamma,\\
	[0,\lambda], & \text{ if } \theta=\gamma,\\
	0, & \text{ if } \theta>\gamma.
\end{cases}$\\\hline
ETP \cite{gao2011feasible}& $\frac{\lambda}{1-\exp(-\gamma)}(1-\exp(-\gamma\theta))$ & $\frac{\lambda\gamma}{1-\exp(-\gamma)}\exp(-\gamma\theta)$\\\hline
Geman \cite{geman1995nonlinear} & $\frac{\lambda\theta}{\theta+\gamma}$ &$\frac{\lambda\gamma}{(\theta+\gamma)^2}$\\\hline
Laplace \cite{trzasko2009highly}& $\lambda(1-\exp(-\frac{\theta}{\gamma}))$ & $\frac{\lambda}{\gamma}\exp(-\frac{\theta}{\gamma})$ \\\hline
\end{tabular}
\vspace{-1em}
\end{table*}

\section{Introduction}
This paper aims to solve the following general nonconvex nonsmooth low-rank minimization problem
\begin{equation}\label{eq_genpro}
\min_{\mathbf{X}\in\mathbb{R}^{m\times n}} F(\X)=\sum_{i=1}^mg_{\lambda}(\sigma_i(\mathbf{X}))+f(\mathbf{X}),
\end{equation}
where $\sigma_i(\X)$ denotes the $i$-th singular value of $\X\in\mathbb{R}^{m\times n}$ (we assume $m\leq n$ in this work). 
The penalty function $g_{\lambda}$ and loss function $f$ satisfy the following assumptions:
\begin{itemize}
\item[\textbf{A1}] $g_\lambda:$ $\mathbb{R}\rightarrow\mathbb{R}^+$ is continuous, concave and monotonically increasing on $[0,\infty)$. It is possibly nonsmooth.
\item[\textbf{A2}] $f$: $\mathbb{R}^{m\times n}\rightarrow\mathbb{R}^+$ is a smooth function of type $C^{1,1}$, i.e., the gradient is Lipschitz continuous,
\begin{equation}
||\nabla f(\mathbf{X})-\nabla f(\mathbf{Y})||_F\leq L(f)||\mathbf{X}-\mathbf{Y}||_F,
\end{equation}
for any $\mathbf{X}, \mathbf{Y}\in\mathbb{R}^{m\times n}$, $L(f)>0$ is called Lipschitz constant of $\nabla f$. $f(\X)$ is possibly nonconvex.
\item[\textbf{A3}] $F(\X)\rightarrow\infty$ iff $||\X||_F\rightarrow\infty$.
\end{itemize}

Many optimization problems in machine learning and computer vision areas fall into the formulation in (\ref{eq_genpro}). As for the choice of $f$, the squared loss $f(\X)=\frac{1}{2}||\mathcal{A}(\X)-\mathbf{b}||_F^2$, with a linear mapping $\mathcal{A}$, is widely used. In this case, the Lipschitz constant of $\nabla f$ is then the spectral radius of $\mathcal{A}^*\mathcal{A}$, i.e., $L(f)=\rho(\mathcal{A}^*\mathcal{A})$, where $\mathcal{A}^*$ is the adjoint operator of $\mathcal{A}$. By choosing $g_\lambda(x)=\lambda x$, $\sum_{i=1}^mg_\lambda(\sigma_i(\X))$ is exactly the nuclear norm $\lambda\sum_{i=1}^m\sigma_i(\X)=\lambda||\X||_*$. Problem (\ref{eq_genpro}) resorts to the well known nuclear norm regularized problem
\begin{equation}\label{pro_nuclearmi}
\min_{\X} \lambda||\X||_*+f(\X).
\end{equation}
If $f(\X)$ is convex, it is the most widely used convex relaxation of the rank minimization problem:
\begin{equation}\label{pro_rankmi}
\min_{\X} \lambda\text{rank}(\X)+f(\X).
\end{equation}
The above low-rank minimization problem arises in many machine learning tasks such as
multiple category classification \cite{amit2007uncovering},
matrix completion \cite{toh2010accelerated}, multi-task learning \cite{argyriou2008convex}, and low-rank representation with squared loss for subspace segmentation \cite{robustlrr}. However, solving problem (\ref{pro_rankmi}) is usually  difficult, or even NP-hard. Most previous works solve the convex problem (\ref{pro_nuclearmi}) instead. It has been proved that under certain incoherence assumptions on the singular values of the matrix, solving the convex nuclear norm regularized problem leads to a near optimal low-rank solution \cite{candes2010power}. However, such assumptions may be violated in real applications. The obtained solution by using nuclear norm may be suboptimal since it is not a perfect approximation of the rank function. A similar phenomenon has been observed in the convex $L_1$-norm and nonconvex $L_0$-norm for sparse vector recovery \cite{candes2008enhancing}.

In order to achieve a better approximation of the $L_0$-norm, many nonconvex surrogate functions of $L_0$-norm have been proposed, including $L_p$-norm \cite{frank1993statistical}, Smoothly Clipped Absolute Deviation (SCAD) \cite{fan2001variable}, Logarithm \cite{friedman2012fast}, Minimax Concave Penalty (MCP) \cite{zhang2010nearly}, Capped $L_1$ \cite{zhang2010analysis}, Exponential-Type Penalty (ETP) \cite{gao2011feasible}, Geman \cite{geman1995nonlinear}, and Laplace \cite{trzasko2009highly}. Table \ref{tab_nonpenlty} tabulates these penalty functions and Figure \ref{fig_nonconfun} visualizes them. One may refer to \cite{gasso2009recovering} for more properties of these penalty functions. Some of these nonconvex penalties have been extended to approximate the rank function, e.g. the Schatten-$p$ norm \cite{IRLSrank}. Another nonconvex surrogate of rank function is the truncated nuclear norm \cite{hu2012fast}.

For nonconvex sparse minimization, several algorithms have been proposed to solve the problem with a nonconvex regularizer.  A common method is DC (Difference of Convex functions) programming  \cite{gasso2009recovering}. It minimizes the nonconvex function $f(\x)-(-g_\lambda(\x))$ based on the assumption that both $f$ and $-g_\lambda$ are convex. In each iteration, DC programming linearizes $-g_\lambda(\x)$ at $\x={\x}^k$, and minimizes the relaxed function as follows
\begin{equation}\label{pro_dc}
{\x}^{k+1}=\arg\min_{\x} f(\x)-(-g_\lambda({\x}^k))-\left\langle\mathbf{v}^k,\x-{\x}^k\right\rangle,
\end{equation}
where ${\mathbf{v}}^k$ is a subgradient of $-g_\lambda(\x)$ at $\x=\x_k$. DC programming may be not very efficient, since it requires some other iterative algorithm to solve (\ref{pro_dc}). Note that the updating rule (\ref{pro_dc}) of DC programming cannot be extended to solve the low-rank problem (\ref{eq_genpro}). The reason is that for concave $g_\lambda$, $-\sum_{i=1}^mg_\lambda(\sigma_i(\X))$ does not guarantee to be convex w.r.t. $\X$. DC programming also fails when $f$ is nonconvex in problem (\ref{eq_genpro}).
%
%
%

%
%


Another solver is to use the proximal gradient algorithm which is originally designed for convex problem \cite{beck2009fast}. It requires computing the proximal operator of $g_\lambda$,
\begin{equation}\label{pro_oper}
P_{g_\lambda}(y)=\arg\min_x g_\lambda(x)+\frac{1}{2}(x-y)^2,
\end{equation}
in each iteration. However, for nonconvex $g_\lambda$, there may not exist a general solver for (\ref{pro_oper}). Even if (\ref{pro_oper}) is solvable, different from convex optimization, $(P_{g_\lambda}(y_1)-P_{g_\lambda}(y_2))(y_1-y_2)\geq0$ does not always hold. Thus we cannot perform $P_{g_\lambda}(\cdot)$ on the singular values of $\Y$ directly for solving
\begin{equation}
P_{g_\lambda}(\Y)=\arg\min_{\X} \sum_{i=1}^m g_\lambda(\sigma_i(\X))+||\X-\Y||_F^2.
\end{equation}
The nonconvexity of $g_\lambda$ makes the nonconvex low-rank minimization problem much more challenging than the nonconvex sparse minimization.

Another related work is the Iteratively Reweighted Least Squares (IRLS) algorihtm. It has been recently extended to handle the nonconvex Schatten-$p$ norm penalty \cite{IRLSrank}. Actually it solves a relaxed smooth problem which may require many iterations to achieve a low-rank solution. It cannot solve the general nonsmooth problem (\ref{eq_genpro}). The alternative updating algorithm in \cite{hu2012fast} minimizes the truncated nuclear norm by using a special property of this penalty. It contains two loops, both of which require computing SVD. Thus it is not very efficient. It cannot be extended to solve the general problem (\ref{eq_genpro}) either.


In this work, all the existing nonconvex surrogate functions of $L_0$-norm are extended on the singular values of a matrix to enhance low-rank recovery. In problem (\ref{eq_genpro}), $g_\lambda$ can be any existing nonconvex penalty function shown in Table \ref{tab_nonpenlty} or any other function which satisfies the assumption (\textbf{A1}). We observe that all the existing nonconvex surrogate functions are concave and monotonically increasing on $[0,\infty)$. Thus their gradients (or supergradients at the nonsmooth points) are nonnegative and monotonically decreasing. Based on this key fact, we propose an Iteratively Reweighted Nuclear Norm (IRNN) algorithm to solve problem (\ref{eq_genpro}). IRNN computes the proximal operator of the weighted nuclear norm, which has a closed form solution due to the nonnegative and monotonically decreasing supergradients. In theory, we prove that IRNN monotonically decreases the objective function value, and any limit point is a stationary point. \textbf{To the best of our knowledge, IRNN is the first work which is able to solve the general problem (\ref{eq_genpro}) with convergence guarantee. Note that for nonconvex optmization, it is usually very difficult to prove that an algorithm converges to stationary points}. At last, we test our algorithm with several nonconvex penalty functions on both synthetic data and real image data to show the effectiveness of the proposed algorithm.

\begin{figure}[!t]
\centering
\includegraphics[width=0.45\textwidth]{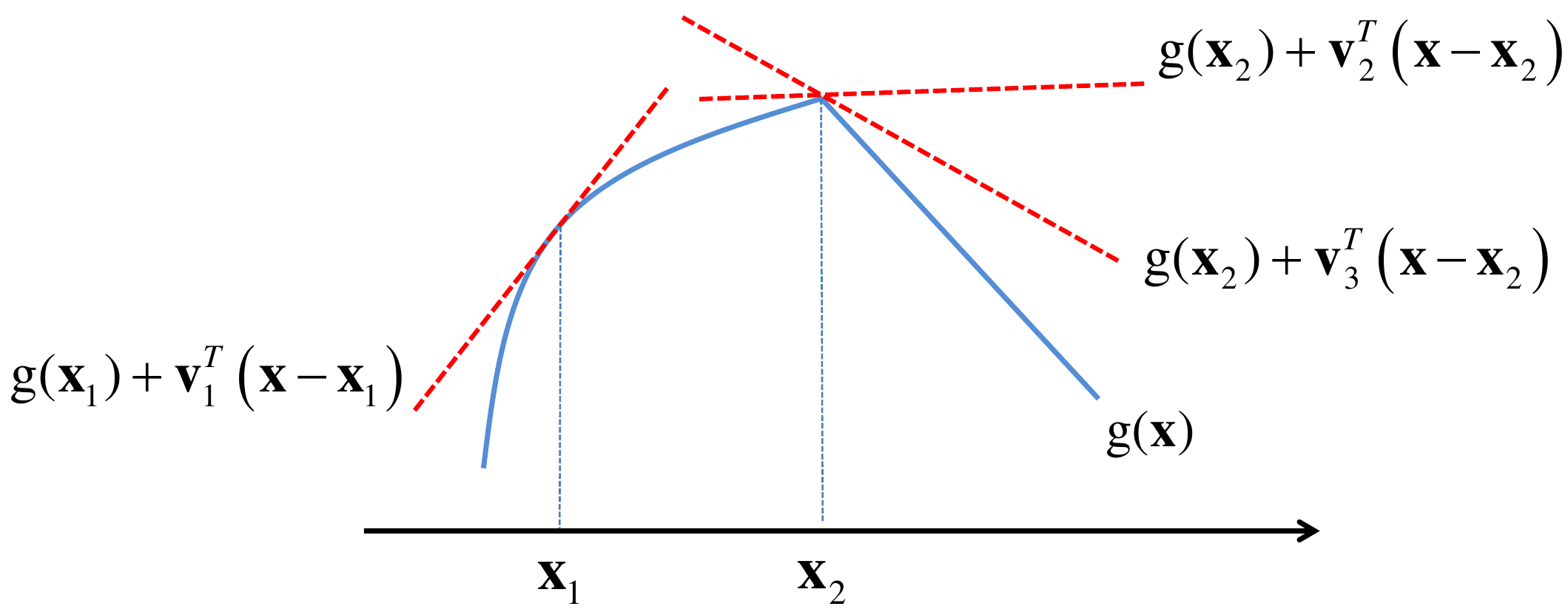}
\caption{\small{Supergraidients of a concave function. $\mathbf{v}_1$ is a supergradient at $\mathbf{x}_1$, and $\mathbf{v}_2$ and $\mathbf{v}_3$ are supergradients at $\mathbf{x}_2$.}}
\label{fig_supergradient}
\vspace{-1em}
\end{figure}

\section{Nonconvex Nonsmooth Low-Rank Minimization}
In this section, we present a general algorithm to solve problem (\ref{eq_genpro}). To handle the case that $g_\lambda$ is nonsmooth, e.g., Capped $L_1$ penalty, we need the concept of supergradient defined on the concave function.
\subsection{Supergradient of a Concave Function}
\label{sec_21}
The subgradient of the convex function is an extension of gradient at a nonsmooth point. Similarly, the supergradient is an extension of gradient of the concave function at a nonsmooth point. If $g(\x)$ is concave and differentiable at $\x$, it is known that
\begin{equation}\label{eq_concavefun}
g(\x)+\langle\nabla g(\x),\y-\x\rangle\geq g(\y).
\end{equation}
If $g(\x)$ is nonsmooth at $\x$, the supergradient extends the gradient at $\x$ inspired by (\ref{eq_concavefun}) \cite{border20011}.
\begin{defi}
Let $g: \mathbb{R}^n\rightarrow\mathbb{R}$ be concave. A vector $\mathbf{v}$ is a supergradient of $g$ at the point $\x\in\mathbb{R}^n$ if for every $\y\in\mathbb{R}^n$, the following inequality holds
\begin{equation}\label{eq_defsuper}
g(\x)+\langle \mathbf{v},\y-\x\rangle\geq g(\y).
\end{equation}
\end{defi}
All supergradients of $g$ at $\x$ are called the superdifferential of $g$ at $\x$, and are denoted as $\partial g(\x)$. If $g$ is differentiable at $\x$, $\nabla g(\x)$ is also a supergradient, i.e., $\partial g(\x)=\{\nabla g(\x)\}$. Figure \ref{fig_supergradient} illustrates the supergradients of a concave function at both differentiable and nondifferentiable points.

For concave $g$, $-g$ is convex, and vice versa. From this fact, we have the following relationship between the supergradient of $g$ and the subgradient of $-g$.
\begin{lem}\label{lem_super_sub}
Let $g(\x)$ be concave and $h(\x)=-g(\x)$. For any $\mathbf{v}\in\partial g(\x)$, $\mathbf{u}=-\mathbf{v}\in\partial h(\x)$, and vice versa.
\end{lem}

The relationship of the supergradient and subgradient shown in Lemma \ref{lem_super_sub} is useful for exploring some properties of the supergradient. It is known that the subdiffierential of a convex function $h$ is a
monotone operator, i.e.,
\begin{equation}\label{eq_monotone}
\langle\mathbf{u}-\mathbf{v},\x	-\y\rangle\geq 0,
\end{equation}
for any $\mathbf{u}\in\partial h(\x)$, $\mathbf{v}\in\partial h(\y)$. The superdifferential of a concave function holds a similar property, which is called antimonotone operator in this work.
\begin{lem}\label{lem_antimo}
The superdifferential of a concave function $g$ is an antimonotone operator, i.e.,
\begin{equation}\label{eq_antimo}
\langle\mathbf{u}-\mathbf{v},\x	-\y\rangle\leq 0,
\end{equation}
for any $\mathbf{u}\in\partial g(\x)$, $\mathbf{v}\in\partial g(\y)$.
\end{lem}
This can be easily proved by Lemma \ref{lem_super_sub} and (\ref{eq_monotone}).

Lemma \ref{lem_antimo} is a key lemma in this work. Supposing that the assumption (\textbf{A1}) holds for $g(x)$, (\ref{eq_antimo}) indicates that
\begin{equation}\label{eq_antimono}
u\geq v, \text{ for any } u\in\partial g(x) \text{ and } v\in\partial g(y),
\end{equation}
when $x\leq y$. That is to say, the supergradient of $g$ is monotonically decreasing on $[0,\infty)$. Table \ref{tab_nonpenlty} shows some usual concave functions and their supergradients. We also visualize them in Figure \ref{fig_nonconfun}. It can be seen that they all satisfy the  assumption (\textbf{A1}). Note that for the $L_p$ penalty, we further define that $\partial g(0)=\infty$. This will not affect our algorithm and convergence analysis as shown latter. The Capped $L_1$ penalty is nonsmooth at $\theta=\gamma$, with the superdifferential $\partial g_\lambda(\gamma)=[0,\lambda]$.

\subsection{Iteratively Reweighted Nuclear Norm}
\label{sec_22}
In this subsection, we show how to solve the general nonconvex and possibly nonsmooth  problem (\ref{eq_genpro}) based on the assumptions (\textbf{A1})-(\textbf{A2}). For simplicity of notation, we denote $\sigma_i=\sigma_i(\X)$ and $\sigma_i^k=\sigma_i(\X^k)$.

Since $g_\lambda$ is concave on $[0,\infty)$, by the definition of the supergradient, we have
\begin{equation}\label{eq_concaveg}
g_\lambda(\sigma_i)\leq g_\lambda(\sigma^k_i)+w_i^k(\sigma_i-\sigma_i^k),
\end{equation}
where
\begin{equation}\label{eq_updatew}
w_i^k\in\partial g_\lambda(\sigma_i^k).
\end{equation}
Since $\sigma_1^k\geq\sigma_2^k\geq\cdots\geq\sigma_m^k\geq0$, by the antimonotone property of supergradient (\ref{eq_antimono}), we have
\begin{equation}\label{eq_incresw}
0\leq w_1^k\leq w_2^k\leq\cdots\leq w_m^k.
\end{equation}
This property is important in our algorithm shown latter. (\ref{eq_concaveg}) motivates us to minimize its right hand side instead of $g_\lambda(\sigma_i)$. Thus we may solve the following relaxed problem
\begin{equation}\label{eq_updatexold}
\begin{split}
{\X}^{k+1}=&\arg\min_{\X} \sum_{i=1}^m g_\lambda(\sigma^k_i)+w_i^k(\sigma_i-\sigma_i^k)+f(\X)\\
=&\arg\min_{\X}\sum_{i=1}^m w_i^k\sigma_i+f(\X).
\end{split}
\end{equation}
It seems that updating ${\X}^{k+1}$ by solving the above weighted nuclear norm problem (\ref{eq_updatexold}) is an extension of the weighted $L_1$-norm problem in IRL1 algorithm \cite{candes2008enhancing} (IRL1 is a special DC programming algorithm). However, the weighted nuclear norm is nonconvex in (\ref{eq_updatexold}) (it is convex if and only if $w_1^k\geq w_2^k\geq\cdots\geq w_m^k\geq0$ \cite{chen2012reduced}), while the weighted $L_1$-norm is convex. Solving the nonconvex problem (\ref{eq_updatexold}) is much more challenging than the convex weighted $L_1$-norm problem. In fact, it is not easier than solving the original problem (\ref{eq_genpro}).

\begin{algorithm}[t]
\caption{Solving problem (\ref{eq_genpro}) by IRNN}
\textbf{Input:} $\mu>L(f)$ - A Lipschitz constant of $\nabla f({\X})$.\\
\textbf{Initialize:} $k=0$, ${\X}^k$, and $w^k_i$, $i=1,\cdots,m$.\\
\textbf{Output:} $X^*$. \\
\textbf{while} not converge \textbf{do}
\begin{enumerate}
  \item Update ${\X}^{k+1}$ by solving problem (\ref{eq_updatexlin}).
  \item Update the weights $w_i^{k+1}$, $i=1,\cdots,m$, by
    \begin{equation}
    w_i^{k+1}\in\partial g_\lambda\left(\sigma_i({\X}^{k+1})\right).
    \end{equation}
\textbf{end while}
\end{enumerate}
\label{alg_lirnn}
\end{algorithm}

Instead of updating ${\X}^{k+1}$ by solving (\ref{eq_updatexold}), we linearize $f(\X)$ at ${\X^k}$ and add a proximal term: 
\begin{equation*}
f({\X})\approx f({\X}^k)+\langle\nabla f({\X}^k),{\X}-{\X}^k\rangle+\frac{\mu}{2}||{\X}-{\X}^k||_F^2,
\end{equation*}
where $\mu> L(f)$. Such a choice of $\mu$ guarantees the convergence of our algorithm as shown latter.   Then we update ${\X}^{k+1}$ by solving
\begin{equation}\label{eq_updatexlin}
\begin{split}
{\X}^{k+1}=&\arg\min_{{\X}} \sum_{i=1}^mw_i^k\sigma_i+f({\X}^k)\\
&+\langle\nabla f({\X}^k),{\X}-{\X}^k\rangle+\frac{\mu}{2}||{\X}-{\X}^k||_F^2\\
=& \arg\min_{{\X}} \sum_{i=1}^mw_i^k\sigma_i+\frac{\mu}{2}\left\|{\X}-\left({\X}^k-\frac{1}{\mu}\nabla f({\X}^k)\right)\right\|_F^2.
\end{split}
\end{equation}
Problem (\ref{eq_updatexlin}) is still nonconvex. Fortunately, it has a closed form solution due to (\ref{eq_incresw}).

\begin{lem}\label{Lem_ineq1}
\cite[Theorem 2.3]{chen2012reduced}
For any $\lambda>0$, ${\Y}\in\mathbb{R}^{m\times n}$ and $0\leq w_1\leq w_2\leq\cdots\leq w_s \ (s=\min(m,n))$, a globally optimal solution to the following problem
\begin{equation}
\min \lambda\sum_{i=1}^sw_i\sigma_i({\X})+\frac{1}{2}||{\X}-{\Y}||_F^2,
\end{equation}
is given by the weighted singular value thresholding 
\begin{equation}
{\X}^*=U\mathcal{S}_{\lambda \bm{w}}(\bm{\Sigma})\bm{V}^T,
\end{equation}
where ${\Y}=\bm{U}\bm{\Sigma}\bm{V}^T$ is the SVD of ${\Y}$, and $\mathcal{S}_{\lambda \bm{w}}(\bm{\Sigma})=\Diag\{(\bm{\Sigma}_{ii}-\lambda w_i)_+\}$.
\end{lem}

It is worth mentioning that for the $L_p$ penalty, if $\sigma_i^k=0$, $w_i^k\in\partial g_\lambda(\sigma_i^k)=\{\infty\}$. By the updating rule of $\X^{k+1}$ in (\ref{eq_updatexlin}), we have $\sigma_i^{k+1}=0$. This guarantees that the rank of the sequence $\{\X^k\}$ is nonincreasing.

Iteratively updating $w_i^k$, $i=1,\cdots,m$, by (\ref{eq_updatew}) and ${\X}^{k+1}$ by (\ref{eq_updatexlin}) leads to the proposed Iteratively Reweighted Nuclear Norm (IRNN) algorithm. The whole procedure of IRNN is shown in Algorithm \ref{alg_lirnn}. If the Lipschitz constant $L(f)$ is not known or computable, the backtracking rule can be used to estimate $\mu$ in each iteration \cite{beck2009fast}.

\section{Convergence Analysis}

In this section, we give the convergence analysis for the IRNN algorithm. We will show that IRNN decreases the objective function value monotonically, and any limit point is a stationary point of problem (\ref{eq_genpro}). We first recall the following well-known and fundamental property for a smooth function in the class $C^{1,1}$.
\begin{lem}\label{Lem_lips}
\cite{bertsekas1995nonlinear,beck2009fast}
Let $f: \mathbb{R}^{m\times n}\rightarrow\mathbb{R}$ be a continuously differentiable function with Lipschitz continuous gradient and Lipschitz constant $L(f)$. Then, for any ${\X}, {\Y}\in\mathbb{R}^{m\times n}$, and $\mu\geq L(f)$,
\begin{equation}
f({\X})\leq f({\Y})+\langle {\X}-{\Y},\nabla f({\Y}) \rangle+\frac{\mu}{2}||{\X}-{\Y}||_F^2.
\end{equation}
\end{lem}
\begin{theo}\label{thm_pro}
Assume that $g_\lambda$ and $f$ in problem (\ref{eq_genpro}) satisfy the assumptions (\textbf{A1})-(\textbf{A2}). The sequence $\{{\X}^k\}$ generated in Algorithm \ref{alg_lirnn} satisfies the following properties:
\begin{enumerate}[(1)]
\item $F({\X}^k)$ is monotonically decreasing. 
Indeed,
\begin{equation*}
F({\X}^k)-F({\X}^{k+1})\geq\frac{\mu-L(f)}{2}||{\X}^k-{\X}^{k+1}||_F^2\geq0;
\end{equation*}
\item $\lim\limits_{k\rightarrow\infty}({\X}^k-{\X}^{k+1})=\bm{0}$;
\item The sequence $\{{\X}^k\}$ is bounded.
\end{enumerate}
\end{theo}
\textbf{Proof.} First, since ${\X}^{k+1}$ is a global solution to problem (\ref{eq_updatexlin}), we get
\begin{equation*}\label{eq_proof1}
\begin{split}
&\sum_{i=1}^mw_i^k\sigma_i^{k+1}+\langle\nabla f({\X}^k),{\X}^{k+1}-{\X}^k\rangle+\frac{\mu}{2}||{\X}^{k+1}-{\X}^k||^2_F\\
\leq&\sum_{i=1}^mw_i^k\sigma_i^{k}+\langle\nabla f({\X}^k),{\X}^{k}-{\X}^k\rangle+\frac{\mu}{2}||{\X}^{k}-{\X}^k||^2_F.
\end{split}
\end{equation*}
It can be rewritten as
\begin{equation}\label{eq_proof4}
\begin{split}
&\langle\nabla f({\X}^k),{\X}^k-{\X}^{k+1}\rangle\\
\geq&-\sum_{i=1}^mw_i^k(\sigma_i^k-\sigma_i^{k+1})+\frac{\mu}{2}||{\X}^k-{\X}^{k+1}||_F^2.
\end{split}
\end{equation}
Second, since the gradient of $f({\X})$ is Lipschitz continuous, by using Lemma \ref{Lem_lips}, we have
\begin{equation}\label{eq_proof5}
\begin{split}
&f({\X}^k)-f({\X}^{k+1})\\
\geq&\langle\nabla f({\X}^k),{\X}^k-{\X}^{k+1}\rangle-\frac{L(f)}{2}||{\X}^k-{\X}^{k+1}||_F^2.
\end{split}
\end{equation}
Third, since $w_i^k\in\partial g_\lambda(\sigma_i^k)$, by the definition of the supergradient, we have
\begin{equation}\label{eq_proof6}
g_\lambda(\sigma_i^k)-g_\lambda(\sigma_i^{k+1})\geq w_i^k(\sigma_i^k-\sigma_i^{k+1}).
\end{equation}
Now, summing (\ref{eq_proof4}), (\ref{eq_proof5}) and (\ref{eq_proof6}) for $i=1,\cdots,m$, together, we obtain
\begin{equation}\label{eq_proof7}
\begin{split}
& F({\X}^k)-F({\X}^{k+1})\\
=&\sum_{i=1}^m\left(g_\lambda(\sigma_i^k)-g_\lambda(\sigma_i^{k+1}) \right)+f({\X}^k)-f({\X}^{k+1}) \\
\geq&\frac{\mu-L(f)}{2}||{\X}^{k+1}-{\X}^k||_F^2\geq0.\\
\end{split}
\end{equation}
Thus $F({\X}^k)$ is monotonically decreasing. Summing all the inequalities in (\ref{eq_proof7}) for $k\geq1$, we get
\begin{equation}\label{eq_proof9}
F({\X}^1)\geq\frac{\mu-L(f)}{2}\sum_{k=1}^\infty||{\X}^{k+1}-{\X}^k||_F^2,
\end{equation}
or equivalently,
\begin{equation}
\sum\limits_{k=1}^\infty||{\X}^k-{\X}^{k+1}||_F^2\leq \frac{2F(\X^1)}{\mu-L(f)}.
\end{equation}
In particular, it implies that $\lim\limits_{k\rightarrow\infty}({\X}^k-{\X}^{k+1})=\bm{0}$. 
The boundedness of $\{{\X}^k\}$ is obtained based on the assumption (\textbf{A3}).
$\hfill\blacksquare$

\begin{theo}\label{thm_con}
Let $\{{\X}^k\}$ be the sequence generated in Algorithm \ref{alg_lirnn}. Then any accumulation point ${\X}^*$ of $\{{\X}^k\}$ is a stationary point of (\ref{eq_genpro}).
\end{theo}
\textbf{Proof.} The sequence $\{{\X}^k\}$ generated in Algorithm \ref{alg_lirnn} is bounded as shown in Theorem \ref{thm_pro}. Thus there exists a matrix ${\X}^*$ and a subsequence $\{{\X}^{k_{j}}\}$ such that $\lim\limits_{j\rightarrow\infty}{\X}^{k_{j}}={\X}^*$. From the fact that $\lim\limits_{k\rightarrow\infty}({\X}^{k}-{\X}^{k+1})=\bm{0}$ in Theorem \ref{thm_pro}, we have $\lim\limits_{j\rightarrow\infty}{\X}^{k_{j}+1}={\X}^*$. Thus $\sigma_i({\X}^{k_j+1})\rightarrow\sigma_i({\X}^*)$ for $i=1,\cdots,m$. By the choice of $w_i^{k_j}\in\partial g_\lambda(\sigma_i({\X}^{k_j}))$ and Lemma \ref{lem_super_sub}, we have $-w_i^{k_j}\in\partial \left(-g_\lambda(\sigma_i({\X}^{k_j}))\right)$. By the upper semi-continuous property of the subdifferential \cite[Proposition 2.1.5]{clarke1983nonsmooth}, there exists $-w_i^*\in\partial \left(-g_\lambda(\sigma_i(\X^*))\right)$ such that $-w_i^{k_j}\rightarrow -w_i^*$. Again by Lemma \ref{lem_super_sub}, $w_i^*\in\partial g_\lambda(\sigma_i(\X^*))$ and $w_i^{k_j}\rightarrow w_i^*$.

Denote $h({\X},\bm{w})=\sum_{i=1}^mw_i\sigma_i({\X})$. Since ${\X}^{k_{j}+1}$ is optimal to problem (\ref{eq_updatexlin}), there exists
$\bm{G}^{k_j+1}\in\partial h({\X}^{k_j+1},\bm{w}^{k_j})$, such that
\begin{equation}\label{eq_proof11}
\bm{G}^{k_j+1}+\nabla f({\X}^{k_j})+\mu({\X}^{k_j+1}-{\X}^{k_j})=\bm{0}.
\end{equation}
Let $j\rightarrow\infty$ in (\ref{eq_proof11}), there exists $\bm{G}^*\in\partial h({\X}^*,\bm{w}^*)$, such that
\begin{equation}\label{eq_proof12}
\bm{0}=\bm{G}^*+\nabla f({\X}^{*})\in\partial F({\X}^*).
\end{equation}
Thus ${\X}^*$ is a stationary point of (\ref{eq_genpro}).  $\hfill\blacksquare$

\section{Extension to Other Problems}
\label{sec_4}
Our proposed IRNN algorithm can solve a more general low-rank minimization problem as follows,
\begin{equation}
\min_{\X} \sum_{i=1}^m g_i(\sigma_i(\X))+f(\X),
\end{equation}
where $g_i$, $i=1,\cdots,m$, are concave, and their supergradients satisfy $0\leq v_1\leq v_2\leq\cdots\leq v_m$, for any $v_i\in\partial g_i(\sigma_i(\X))$, $i=1,\cdots,m$. The truncated nuclear norm $||\X||_r=\sum_{i=r+1}^m\sigma_i(\X)$ \cite{hu2012fast} satisfies the above assumption. Indeed, $||\X||_r=\sum_{i=1}^mg_i(\sigma_i(\X))$ by letting
\begin{equation}
g_i(x)=\begin{cases} 0, & i=1,\cdots,r,\\ x, & i=r+1,\cdots,m.\end{cases}
\end{equation}
Their supergradients are
\begin{equation}
\partial g_i(x)=\begin{cases} 0, & i=1,\cdots,r,\\ 1, & i=r+1,\cdots,m.\end{cases}
\end{equation}
The convergence results in Theorem \ref{thm_pro} and \ref{thm_con} also hold since (\ref{eq_proof6}) holds for each $g_i$. Compared with the alternating updating algorithms in \cite{hu2012fast}, which require double loops, our IRNN algorithm will be more efficient and with stronger convergence guarantee.

More generally, IRNN can solve the following problem
\begin{equation}\label{pro_stru}
\min_{\X} \sum_{i=1}^m g(h(\sigma_i(\X)))+f(\X),
\end{equation}
when $g(y)$ is concave, and the following problem
\begin{equation}
\min_{\X } w_ih(\sigma_i(\X))+||\X-\mathbf{Y}||_F^2,
\end{equation}
can be cheaply solved. An interesting application of (\ref{pro_stru}) is to extend the group sparsity on the singular values. By dividing the singular values into $k$ groups, i.e., $G_1=\{1,\cdots,r_1\}$, $G_2=\{r_1+1,\cdots,r_1+r_2-1\}$, $\cdots$, $G_k=\{\sum_i^{k-1}r_i+1,\cdots,m\}$, where $\sum_ir_i=m$, we can define the group sparsity on the singular values as $||\X||_{2,g}=\sum_{i=1}^kg(||\mathbf{\sigma}_{G_i}||_2)$. This is exactly the first term in (\ref{pro_stru}) by letting $h$ be the $L_2$-norm of a vector. $g$ can be nonconvex functions satisfying the assumption (\textbf{A1}) or specially the convex absolute function.
\begin{figure}
	\begin{subfigure}[b]{0.233\textwidth}
		\centering
        \includegraphics[width=\textwidth]{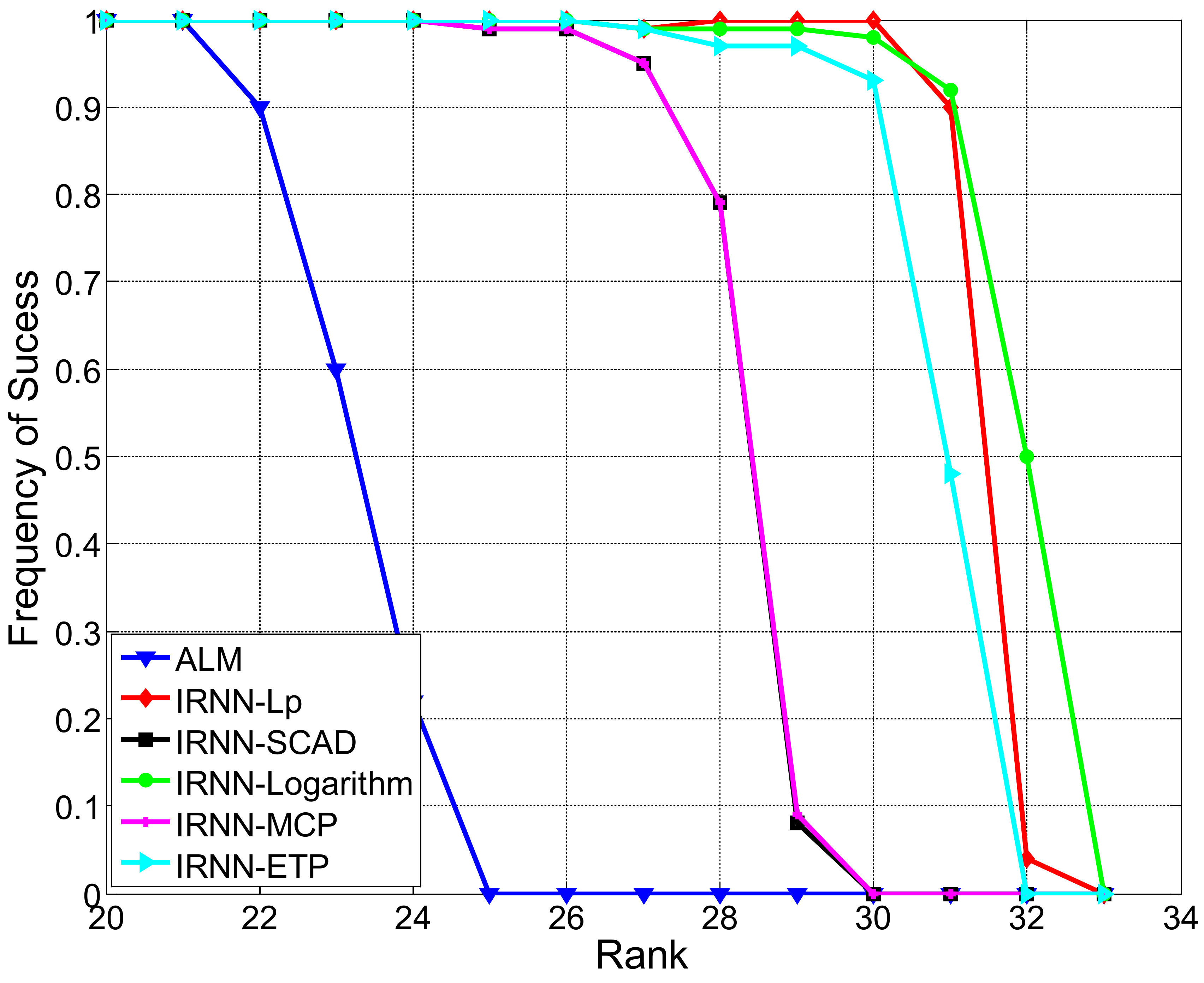}
        \caption{random data without noise}
        \label{fig_randrecov_noiseless}
    \end{subfigure}
    \begin{subfigure}[b]{0.238\textwidth}
		\centering
        \includegraphics[width=\textwidth]{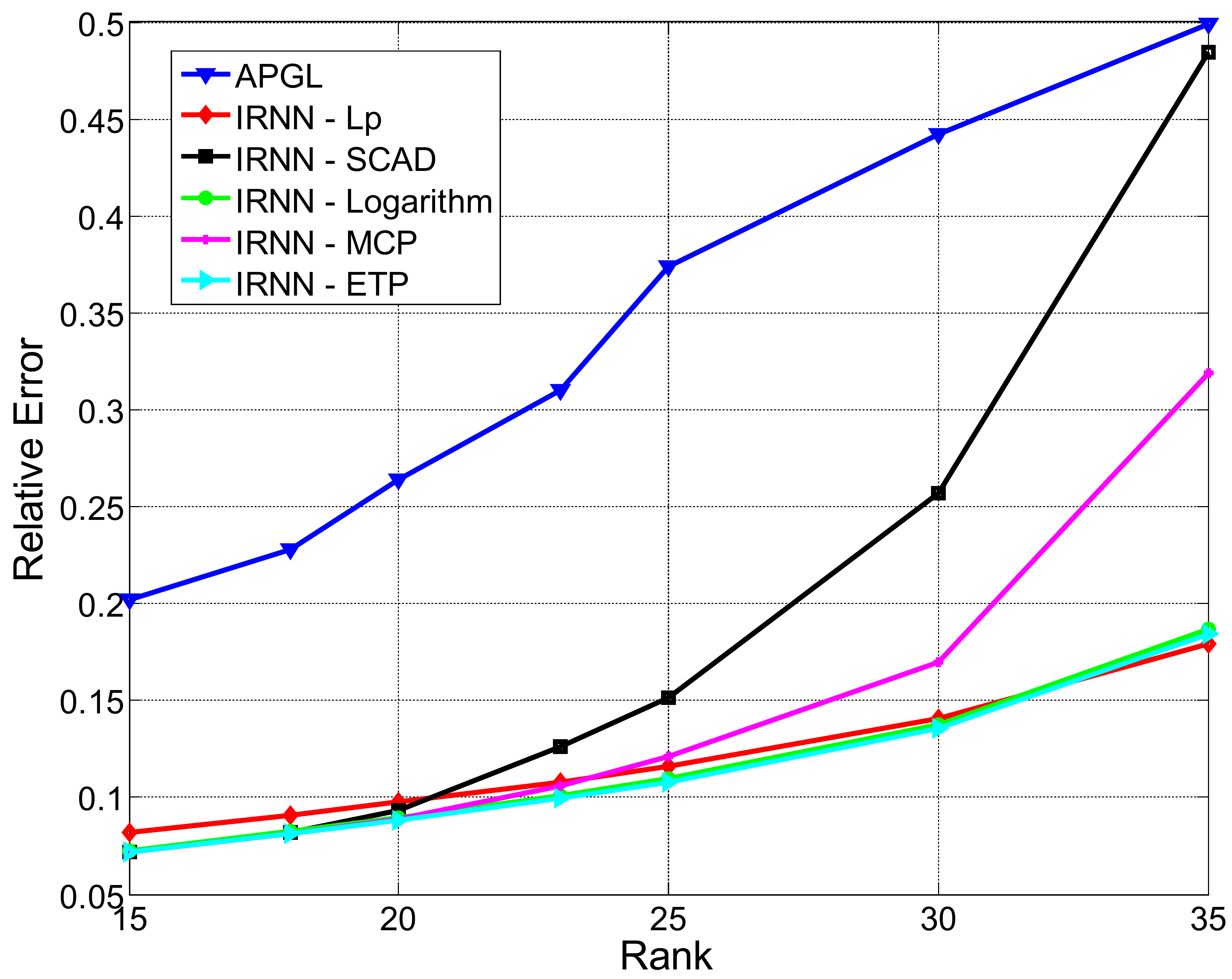}
        \caption{random data with noise}
        \label{fig_randrecov_noise}
    \end{subfigure}
    \caption{\small{Comparison of matrix recovery on (a) random data without noise, and (b) random data with noise.}}\label{fig_randrecov}
    \vspace{-1.5em}
\end{figure}

\begin{figure*}[!t]
\centering %
\includegraphics[width=1\textwidth]{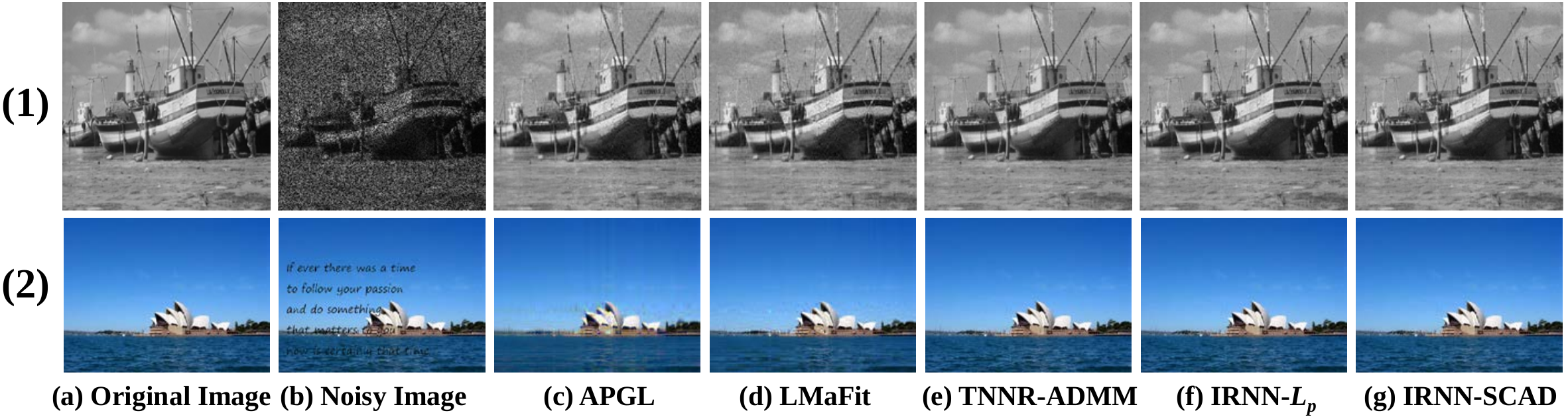}
\caption{{Comparison of image recovery by using different matrix completion algorithms. (a) Original image. (b) Image with Gaussian noise and text. (c)-(g) Recovered images by APGL, LMaFit, TNNR-ADMM, IRNN-$L_p$, and IRNN-SCAD, respectively. \textbf{Best viewed in $\times 2$ sized color pdf file.}}}\label{fig_imagerecovery}
\vspace{-1.8em}
\end{figure*}
%


\section{Experiments}

In this section, we present several experiments on both synthetic data and real images to validate the effectiveness of the IRNN algorithm. We test our algorithm on the matrix completion problem
\begin{equation}\label{pro_mc}
\min_{\X} \sum_{i=1}^mg_\lambda(\sigma_i(\X))+\frac{1}{2}||\mathcal{P}_{\Omega}(\X-\M)||_F^2,
\end{equation}
where $\Omega$ is the set of indices of samples, and $\mathcal{P}_\Omega: \mathbb{R}^{m\times n}\rightarrow\mathbb{R}^{m\times n}$ is a linear operator that keeps the entries in $\Omega$ unchanged and those outside $\Omega$ zeros. The gradient of squared loss function in (\ref{pro_mc}) is Lipschitz continuous, with a Lipschitz constant $L(f)=1$. We set $\mu=1.1$ in Algorithm \ref{alg_lirnn}. For the choice of $g_\lambda$, we test all the penalty functions listed in Table \ref{tab_nonpenlty} except for Capped $L_1$ and Geman, since we find that their recovery performances are sensitive to the choices of $\gamma$ and $\lambda$ in different cases. For the choice of $\lambda$ in IRNN, we use a continuation technique to enhance the low-rank matrix recovery. The initial value of $\lambda$ is set to a larger value $\lambda_0$, and dynamically decreased by $\lambda=\eta^k\lambda_0$ with $\eta<1$. It is stopped till reaching a predefined target $\lambda_t$. $\X$ is initialized as a zero matrix. For the choice of parameters (e.g., $p$ and $\gamma$) in the nonconvex penalty functions, we search it from a candidate set and use the one which obtains good performance in most cases \footnote{Code of IRNN: \scriptsize{\url{https://sites.google.com/site/canyilu/}.}}.

\subsection{Low-Rank Matrix Recovery}
We first compare our nonconvex IRNN algorithm with state-of-the-art convex algorithms on synthetic data. We conduct two experiments. One is for the observed matrix $\M$ without noise, and the other one is for $\M$ with noise.


For the noise free case, we generate the rank $r$ matrix $\M$ as $\M_L\M_R$, where $\M_L\in\mathbb{R}^{150\times r}$, and $\M_R\in\mathbb{R}^{r\times 150}$ are generated by the Matlab command \mcode{randn}. $50\%$ elements of $\M$ are missing uniformly at random. We compare our algorithm with Augmented Lagrange Multiplier (ALM) \footnote{Code: \scriptsize{\url{http://perception.csl.illinois.edu/matrix-rank/sample_code.html}.}} \cite{ALMlin} which solves the noise free problem


\begin{equation}
\min_{\X} ||\X||_* \ \text{ s.t. } \ \mathcal{P}_\Omega(\X)=\mathcal{P}_\Omega(\M).
\end{equation}
For this task, we set $\lambda_0=||\mathcal{P}_\Omega(\M)||_\infty$, $\lambda_t=10^{-5}\lambda_0$, and $\eta=0.7$ in IRNN, and stop the algorithm when $||\mathcal{P}_\Omega(\X-\M)||_F\leq 10^{-5}$. For ALM, we use the default parameters in the released codes. We evaluate the recovery performance by the Relative Error defined as $||\hat{\X}-\M||_F/||\M||_F,$
where $\hat{\X}$ is the recovered solution by a certain algorithm. If the Relative Error is smaller than $10^{-3}$, $\hat{\X}$ is regarded as a successful recovery of $\M$. We repeat the experiments 100 times with the underlying rank $r$ varying from 20 to 33 for each algorithm. The frequency of success is plotted in Figure \ref{fig_randrecov_noiseless}. The legend IRNN-$L_p$ in Figure \ref{fig_randrecov_noiseless} denotes the $L_p$ penalty function used in problem (\ref{eq_genpro}) and solved by our proposed IRNN algorithm. It can be seen that IRNN with all the nonconvex penalty functions achieves much better recovery performance than the convex ALM algorithm. This is because the nonconvex penalty functions approximate the rank function better than the convex nuclear norm.

For the noisy case, the data are generated by $\mathcal{P}_\Omega(\M)=\mathcal{P}_\Omega(\M_L\M_R)$+0.1$\times$\mcode{randn}. We compare our algorithm with convex  Accelerated Proximal Gradient with Line search (APGL) \footnote{Code: \scriptsize{\url{http://www.math.nus.edu.sg/~mattohkc/NNLS.html}.}} \cite{toh2010accelerated} which solves the noisy problem
\begin{equation}
\min_{\X} \lambda||\X||_*+\frac{1}{2}||\mathcal{P}_\Omega(\X)-\mathcal{P}_\Omega(\M)||_F^2.
\end{equation}
For this task, we set $\lambda_0=10||\mathcal{P}_\Omega(\M)||_\infty$, and $\lambda_t=0.1\lambda_0$ in IRNN. All the chosen algorithms are run 100 times with the underlying rank $r$ lying between 15 and 35. The relative errors can be ranging for each test, and the mean errors by different methods are plotted in Figure \ref{fig_randrecov_noise}. It can be seen that IRNN for the nonconvex penalty outperforms the convex APGL for the noisy case. Note that we cannot conclude from Figure \ref{fig_randrecov} that IRNN with $L_p$, Logarithm and ETP penalty functions always perform better than SCAD and MCP, since the obtained solutions are not globally optimal.

\subsection{Application to Image Recovery}

In this section, we apply matrix completion for image recovery. As shown in Figure \ref{fig_imagerecovery}, the real image may be corrupted by different types of noises, e.g., Gaussian noise or unrelated text. Usually the real images are not of low-rank, but the top singular values dominate the main information \cite{hu2012fast}. Thus the corrupted image can be recovered by low-rank approximation. For color images which have three channels, we simply apply matrix completion for each channel independently. The well known Peak Signal-to-Noise Ratio (PSNR) is employed to evaluate the recovery performance. We compare IRNN with some other matrix completion algorithms which have been applied for this task, including APGL, Low-Rank Matrix Fitting (LMaFit) \footnote{Code: \scriptsize{\url{http://lmafit.blogs.rice.edu/}.}}. \cite{wen2012solving} and Truncated Nuclear Norm Regularization (TNNR)  \cite{hu2012fast}.
We use the solver based on ADMM to solve a subproblem of TNNR in the released codes (denoted as TNNR-ADMM) \footnote{Code:  \scriptsize{\url{https://sites.google.com/site/zjuyaohu/}.}}. We try to tune the parameters to be optimal of the chosen algorithms and report the best result.

\begin{figure}[!t]
\centering
\includegraphics[width=0.5\textwidth]{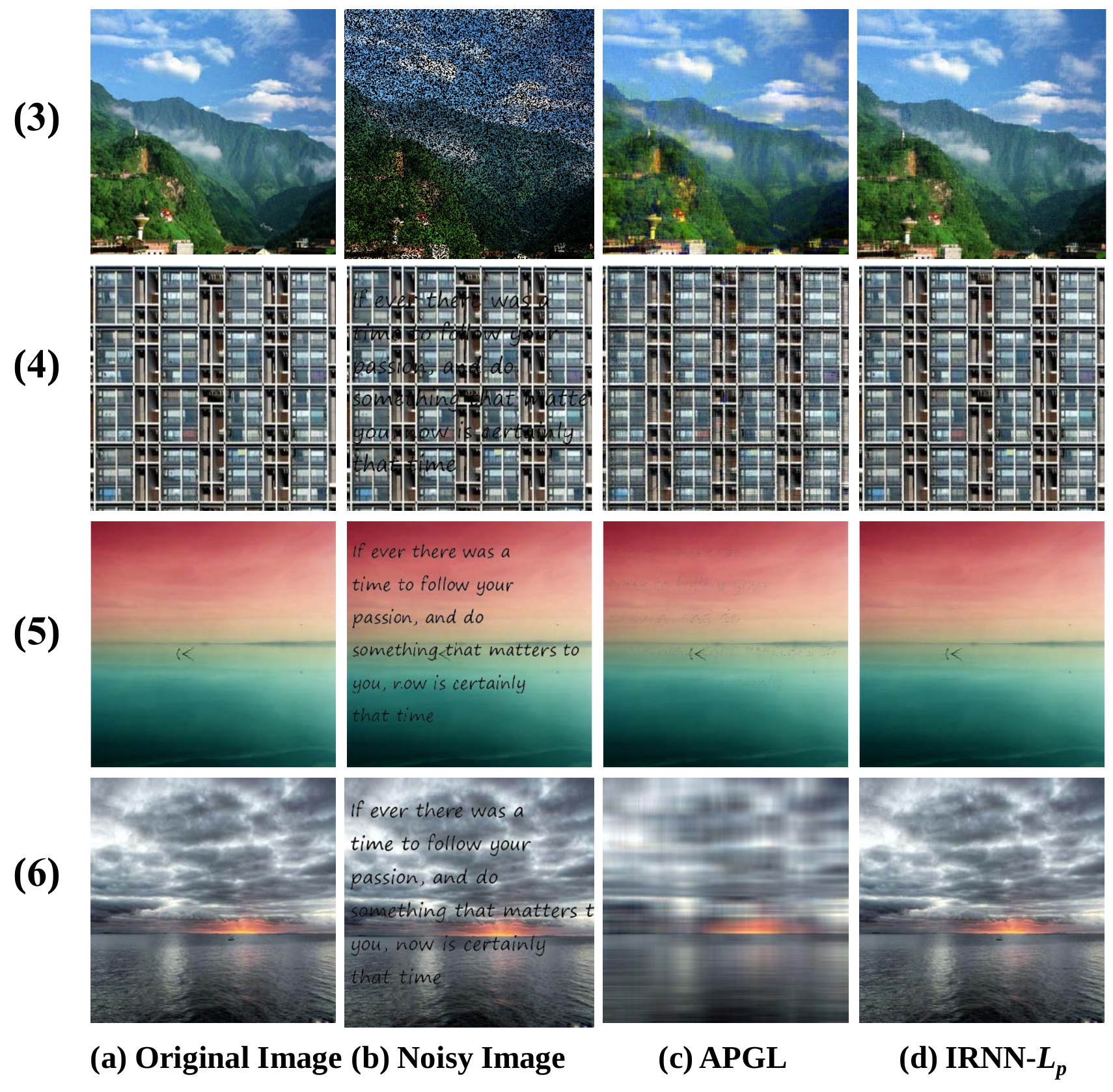}
\caption{\small{Comparison of image recovery on more images. (a) Original images. (b) Images with noises. Recovered images by (c) APGL, and (d) IRNN-$L_p$. \textbf{Best viewed in $\times 2$ sized color pdf file.}} }\label{fig_imagerecovery2}
\vspace{-1em}
\end{figure}

\begin{figure}
	\centering
    \includegraphics[width=0.46\textwidth]{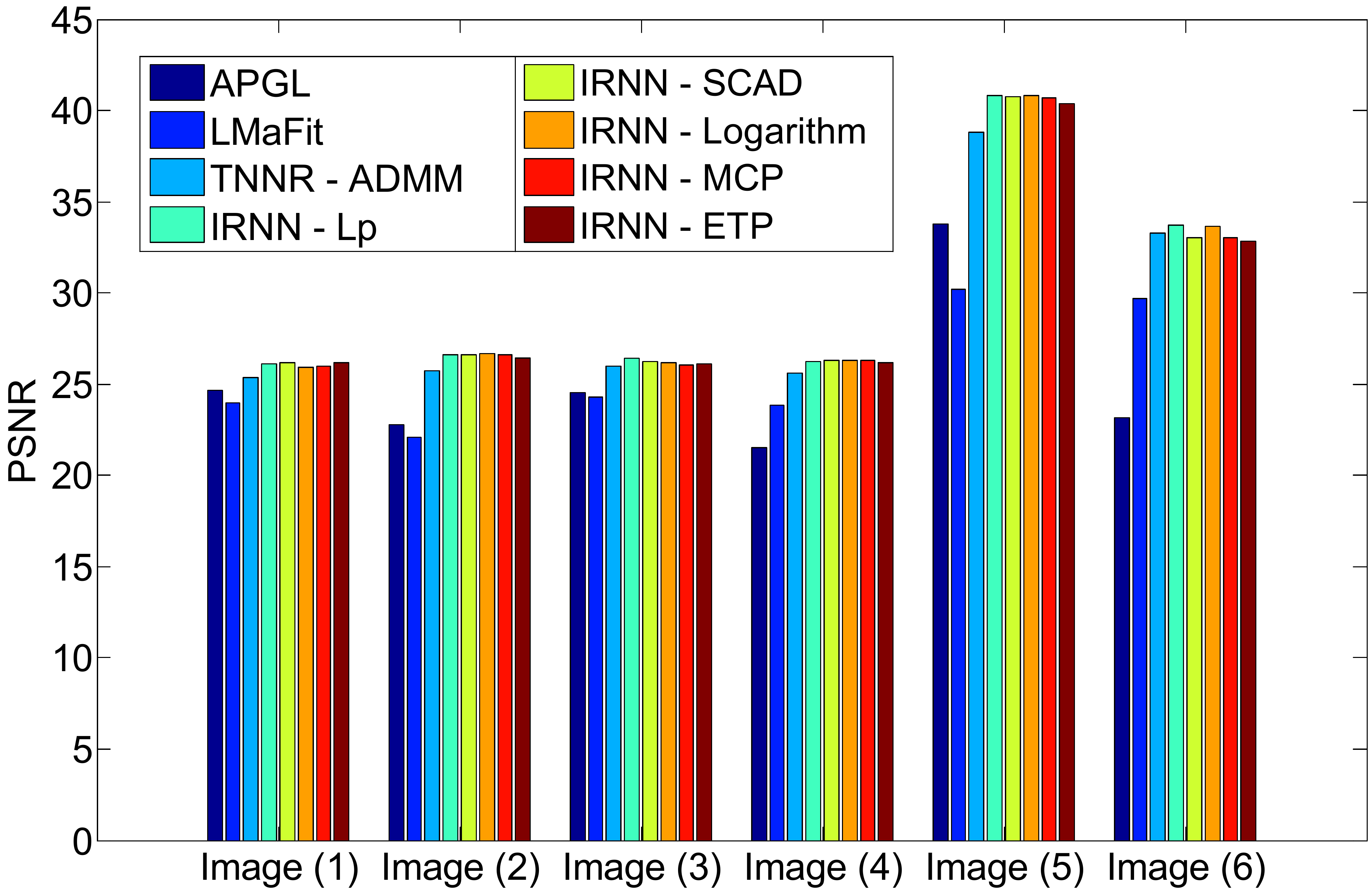}
    \caption{\small{Comparison of the PSNR values by different matrix completion algorithms.}}
    \label{fig_randrecov_psnr}
\vspace{-1em}
\end{figure}

In our test, we consider two types of noises on the real images. The first one replaces $50\%$ of pixels with random values (sample image (1) in Figure \ref{fig_imagerecovery} (b)). The other one adds some unrelated texts on the image (sample image (2) in Figure \ref{fig_imagerecovery} (b)). Figure \ref{fig_imagerecovery} (c)-(g) show the recovered images by different methods. It can be observed that our IRNN method with different penalty functions achieves much better recovery performance than APGL and LMaFit. Only the results by IRNN-$L_p$ and IRNN-SCAD are plotted due to the limit of space. We further test on more images and plot the results in Figure \ref{fig_imagerecovery2}. Figure \ref{fig_randrecov_psnr} shows the PSNR values of different methods on all the test images. It can be seen that IRNN with all the evaluated nonconvex functions achieves higher PSNR values, which verifies that the nonconvex penalty functions are effective in this situation. The nonconvex truncated nuclear norm is close to our methods, but its running time is 3$\sim$5 times of that for ours.
\section{Conclusions and Future Work}
In this work, the nonconvex surrogate functions of $L_0$-norm are extended on the singular values to approximate the rank function. It is observed that all the existing nonconvex surrogate functions are concave and monotonically increasing on $[0,\infty)$. Then a general solver IRNN is proposed to solve problem (\ref{eq_genpro}) with such penalties. IRNN is the first algorithm which is able to solve the general nonconvex low-rank minimization problem (\ref{eq_genpro}) with convergence guarantee. The nonconvex penalty can be nonsmooth by using the supergradient at the nonsmooth point. In theory, we proved that any limit point is a local minimum. Experiments on both synthetic data and real images demonstrated that IRNN usually outperforms the state-of-the-art convex algorithms. An interesting future work is to solve the nonconvex low-rank minimization problem with affine constraint. A possible way is to combine IRNN with Alternating Direction Method of Multiplier (ADMM).

\section*{Acknowledgements}

This research is supported by the Singapore National Research Foundation under its International Research Centre @Singapore Funding Initiative and administered by the IDM Programme Office. Z. Lin is supported by NSF of China (Grant nos. 61272341, 61231002, and 61121002) and MSRA.

{ \scriptsize
\bibliographystyle{ieee}
\bibliography{IRNN}
}
\end{document}